\documentclass[journal]{IEEEtran}

\usepackage{ifpdf}
\usepackage{graphicx}
\usepackage{amsmath}
\usepackage{amssymb}

\usepackage{algorithmic}
\usepackage{array}
\usepackage{url}

\usepackage{times}

\usepackage{multirow}
\usepackage{booktabs}
\usepackage{balance}
\usepackage{CJK}
\usepackage[ruled,vlined]{algorithm2e}
\usepackage{color}

\newcommand{\argmin}{\mathop{\mathrm{arg min}}}

\newcommand{\vb}{{\bf b}}
\newcommand{\vc}{{\bf c}}

\newcommand{\vx}{{\bf x}}

\newcommand{\vH}{{\bf H}}
\newcommand{\vG}{{\bf G}}

\newcommand{\vW}{{\bf W}}

\newcommand{\vS}{{\bf S}}

\newcommand{\vX}{{\bf X}}

\begin{document}

\title{Deep Feature Learning via Structured Graph Laplacian Embedding for Person Re-Identification}

\author{De Cheng,
        Yihong~Gong,~\IEEEmembership{Senior~Member,~IEEE},
        Zhihui~Li, \\
        Weiwei~Shi,
        Alexander G. Hauptmann,
        and~Nanning~Zheng,~\IEEEmembership{Fellow,~IEEE}
\thanks{The authors De Cheng, Yihong Gong, Weiwei Shi and Nanning Zheng are with the Institute of Artificial Intelligence and Robotics,
Xi'an Jiaotong University, Xi¨an 710049, China (e-mail:  chengde19881214@stu.xjtu.edu.cn; ygong@mail.xjtu.edu.cn; shiweiwei.math@stu.xjtu.edu.cn;
 nnzheng@mail.xjtu.edu.cn), Zhihui Li is with Beijing Etrol Technologies Co., Ltd, and Alexander G. Hauptmann is with the Carnegie Mellon University, USA.}
}

\markboth{IEEE TRANSACTIONS ON XXXXXXXXXX}%
{Shi \MakeLowercase{\textit{et al.}}: Deep Feature Learning via Structured Graph Laplacian Embedding for Person Re-Identification}


\maketitle

\begin{abstract}
Learning the distance metric between pairs of examples is of great importance for visual recognition, especially for person re-identification (Re-Id). Recently, the contrastive and triplet loss are proposed to enhance the discriminative power of the deeply learned features, and have achieved remarkable success. As can be seen, either the contrastive or triplet loss is just one special case of the Euclidean distance relationships among these training samples.
Therefore, we propose a structured graph Laplacian embedding algorithm, which can formulate all these structured distance relationships into the graph Laplacian form. The proposed method can take full advantages of the structured distance relationships among these training samples, with the constructed complete graph. Besides, this formulation makes our method easy-to-implement and super-effective.
When embedding the proposed algorithm with the softmax loss for the CNN training, our method can obtain much more robust and discriminative deep features with inter-personal dispersion and intra-personal compactness, which is essential to person Re-Id.
We illustrate the effectiveness of our proposed method on top of three popular networks, namely AlexNet\cite{krizhevsky2012imagenet}, DGDNet\cite{xiao2016learning} and ResNet50\cite{he2015deep}, on recent four widely used Re-Id benchmark datasets. Our proposed method achieves state-of-the-art performances.
\end{abstract}


%
\IEEEpeerreviewmaketitle

\section{Introduction}\label{sec:introduction}

\begin{figure}[t]
 \centering
 \renewcommand{\arraystretch}{.9}
 \renewcommand{\tabcolsep}{.5mm}
 \begin{tabular}{c}
 \includegraphics[width=8cm]{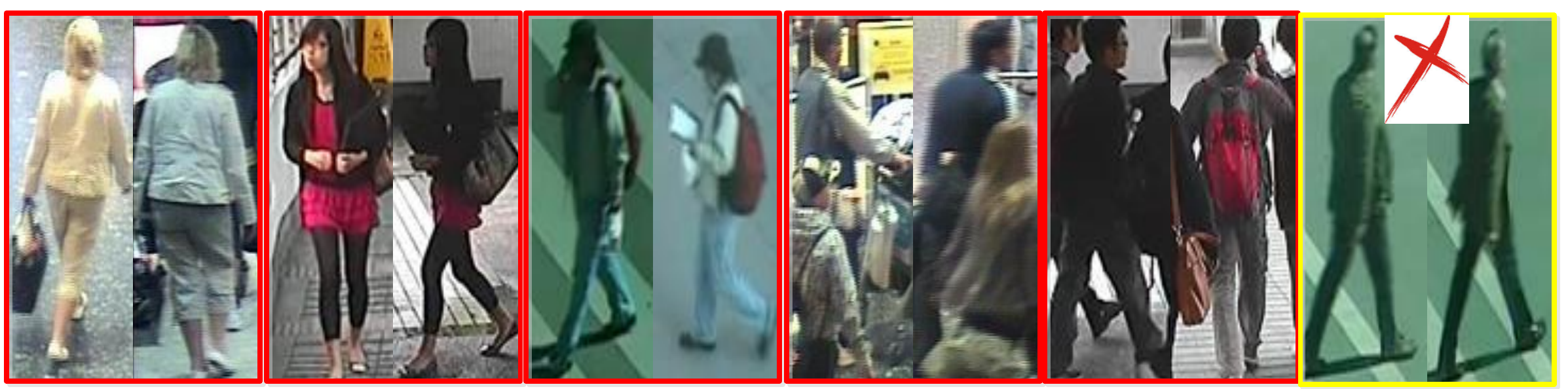}\\
 \includegraphics[width=8cm]{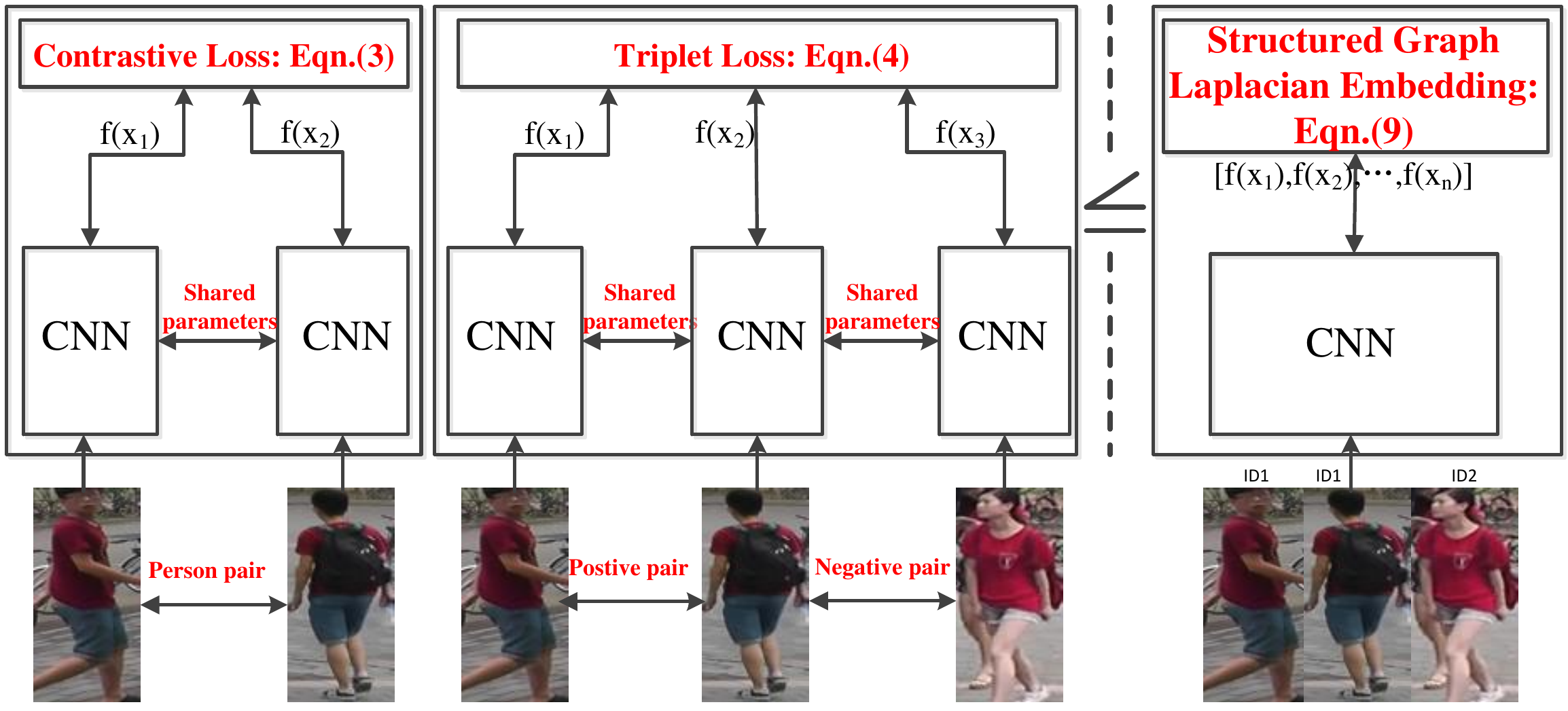}\\
 \end{tabular}
 \caption{\footnotesize{The top row illustrates some examples of different types of challenging difficulties for person Re-Id.
 The bottom row illustrates our proposed method. Compared with the traditional contrastive and triplet loss, the proposed structured graph Laplacian embedding can take full advantage of the Euclidean distance relationships among the training samples by constructing a complete graph, which is easy-to-implement and effective.
 }}
 \label{achitecture}
 \end{figure}
Person Re-Id, which aims at searching pedestrians over a set of  non-overlapping camera views, is an important and challenging task in the computer vision community~\cite{farenzena2010person,gray2008viewpoint,YingZhang2016person,zheng2011person,zheng2016person}. It provides support for video surveillance, and thus saving human labor to search for a certain person, and also helps improve pedestrian tracking among different cameras.
It remains an unsolved problem due to the following reasons: large variances of each person's appearances and poses, illumination changes, different image resolutions and occlusions, as well as similar appearance shared by different persons~\cite{xiao2016cross,cheng2016person}. The top row in Figure~\ref{achitecture} show some examples of aforementioned situations.

Recently, state-of-the-art results on almost all  person Re-Id benchmark datasets are achieved by the deep convolutional neural network (CNN) based methods, which learn the discriminative features and similarity metric simultaneously~\cite{ahmed2015improved,ding2015deep,li2014deepreid,xiao2016learning,yi2014deep,lin2015bilinear}. One of the main advantages of these discriminatively trained network models is that they learn feature representations with the semantically meaningful metric embedding, which aims to reduce the intra-personal variations while increasing the inter-personal differences. Among them, the contrastive~\cite{yi2014deep,geng2016deep,bromley1993signature,ahmed2015improved,yi2014deep}, 
triplet loss ~\cite{cheng2016person,ding2015deep,schroff2015facenet,song2015deep,you2016top}, or their combinations  are widely used~\cite{wen2016discriminative}.

The existing contrastive or triplet related approaches~\cite{cheng2016person,ding2015deep,schroff2015facenet,yi2014deep,geng2016deep} first randomly generate sample pairs or triplets to construct the training batches, then compute the loss with the sample pairs or triplets in the batch, where the loss is the sample pairs' distance comparison, and finally use stochastic gradient decent method  to optimize the network parameters.
As can be seen, both the traditional contrastive and triplet loss are special cases of the Euclidean distance relationships among these training samples. Based on this point, we tend to formulate all these structured distance relationships into the graph Laplacian form. By constructing one complete graph, all these distances between sample pairs can be easily constructed, and the structure relationships among these distances can be reflected by the weight matrix for this complete graph.
Most importantly, this step enables the proposed method to take full advantages of the cluster structure and distance relationships in the training batches at once, which can help to learn more discriminative deep feature embedding.
Specifically, we construct one complete graph with $N$ nodes, and each node corresponds to one sample vector in the batch.
Then we obtain $N \times N$ edges corresponding to the sample pair distances. The weight matrix for the graph edges are constructed by the predefined relationships in the training batches, such as the contrastive or triplet loss. Consequently, the proposed structured graph Laplacian embedding algorithm can take full advantages of all possible pairwise or triplet examples' distance structure relationships within the training batches, by using only one network branch. Besides, this also results in another merit, where neither extra complex data expansion nor redundant memory is needed to fit the traditional siamese or triplet networks.



Since our proposed structured graph Laplacian embedding enjoys the same requirement as the softmax loss and needs no complex data recombination, training the CNN model under the joint supervision of softmax loss and structured graph Laplacian embedding is easy-to-implement. Whereas the softmax loss can enlarge the inter-personal variations among different individuals, and the structured graph Laplacian loss targets more directly on the learning objective of the intra-personal compactness. Therefore, this joint supervision to train the CNN model can help obtain much more robust and discriminative deep features with inter-personal dispersion and intra-personal compactness, which are essential to person Re-Id task. Extensive experiments are conducted  on top of the AlexNet\cite{krizhevsky2012imagenet}, ResNet50\cite{he2015deep} and one reduced GoogLeNet~\cite{szegedy2015going} (denoted as DGDNet\cite{xiao2016learning}) models, and the experimental results on several widely used person Re-Id benchmark datasets show the effectiveness of our proposed method. Moreover, we give a deep insight of three widely used classification deep CNN models on the person Re-Id task, which might be useful for the future work to address the Re-Id problems.

To summarize, the contributions of this paper are three folds:
\begin{itemize}
  \item To the best of our knowledge, we are the first to formulate the structured distance relationships into the graph Laplacian form for deep feature learning, and the proposed structured graph Laplacian embedding  can take full advantages of the Euclidean distance relationships among the training samples in the batch.
  \item By embedding the structured graph Laplacian with  softmax loss for CNN training, our proposed method can obtain much more discriminative deep features with inter-personal dispersion and intra-personal compactness.
  \item Extensive experiments on top of three popular deep models (AlexNet, ResNet50 and DGDNet), demonstrate the effectiveness of our method, and we achieve state-of-the-art performances.
\end{itemize}


\section{Related Work}\label{sec_related}

Our work is related to three lines of active research: 1) Metric learning for person Re-Id, 2) Deep metric learning with CNNs, 3) Joint learning with softmax loss for Re-Id. This section reviews representative works for each category.

\textbf{Metric learning for person Re-Id:} A large number of metric learning and ranking algorithms have been applied to the person Re-Id problem~\cite{xiong2014person,paisitkriangkrai2015learning}. The basic idea behind metric learning is to find a mapping function from feature space to the distance space with certain merits, such as the feature representations  from the  same person being closer than those from different ones~\cite{cheng2016person}. These metric learning methods mainly include the Mahalanobias metric learning~\cite{kostinger2012large}, local fisher analysis ~\cite{weinberger2009distance}, large margin nearest neighbour~\cite{weinberger2009distance}, locally adaptive decision functions~\cite{li2013learning}, attribute based consistent matching~\cite{khamis2014joint}, the polynomial kernel method~\cite{chen2016similarity} and the XQDA algorithm~\cite{liao2015person}.


\begin{figure*}
  \centering
  \includegraphics[width=18cm, height=3.5cm]{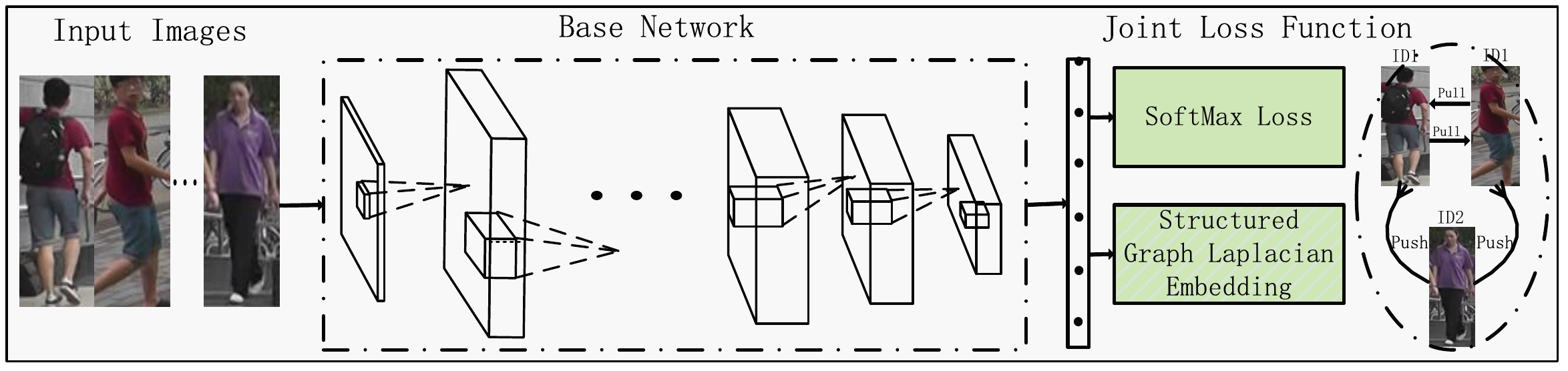}\\
  \caption{\footnotesize{Illustration of our proposed training framework for person Re-Id. This framework is under the joint supervision of softmax loss and the proposed structured graph Laplacian embedding. The softmax loss can enlarge the inter-personal variations, while the proposed algorithm can help reduce the intra-personal variances. We illustrate the effectiveness of our proposed method on top of three base networks: AlexNet, ResNet50 and DGDNet.}}\label{generaFramework}
\end{figure*}

\textbf{Deep metric learning with CNNs:}
Deep learning  based methods~\cite{girshick2014rich,krizhevsky2012imagenet,sun2014deep,he2015deep} have achieved great success in various computer vision tasks,
and it is worth noting that recent state-of-the-art performances on almost all widely used person Re-Id benchmark datasets are obtained by the deep CNN models~\cite{xiao2016cross,xiao2016learning,wang2016joint}.
Among them, many representative works cast the person Re-Id task as a deep metric learning problem and employ pairwise contrastive~\cite{hadsell2006dimensionality,chopra2005learning,hadsell2006dimensionality,chopra2005learning} or verification loss~\cite{bromley1993signature,ahmed2015improved,yi2014deep,shi2015constrained,ustinova2015multiregion,varior2016gated,varior2016siamese}, or triplet ranking loss~\cite{chen2016multi,schroff2015facenet,weinberger2009distance,cheng2016person,ding2015deep}, or their combinations and extensions~\cite{hoffer2015deep,wang2016joint}, to train the networks. While for the network architecture, existing models~\cite{chen2016multi,yi2014deep,ahmed2015improved,wang2016joint,cheng2016person,ding2015deep} differ significantly according to their different training objectives. Their overall network architectures are always with either two or three branches based on the pairwise or triplet loss, respectively.
While in each branch, the network structures differ in the convolution/pooling layers in order to consider the input images' size, aspect ratio of detected person images, and some parts-based properties~\cite{cheng2016person,chen2016multi,yi2014deep}.
In contrast to the aforementioned approaches, our proposed method has formulated the traditional contrastive and triplet loss into the ranking graph Laplacian form, which can take full advantages of the ranking information in the training  batch with only one network branch.


\textbf{Joint learning with the identification loss:} Recently, Xiao~\cite{xiao2016learning} formulated the person Re-Id problem as a conventional classification problem. They adopted  the softmax loss as the identification term, to train one reduced GoogleNet model and achieved superior performance on most person Re-Id benchmark datasets. This indicates that, using the identification loss on top of the deeper network architectures with large datasets, can help achieve good performances. Thus, there are many works aiming to combine softmax loss with other contrastive or verification loss to further improve the Re-Id performances~\cite{xiao2016cross,zheng2016discriminatively}. Combining the two losses aims to exploit both strengths: the classification loss pulls different classes apart and the verification or contrastive loss can reduce the intra-class variations. All of the recently proposed joint learning methods for person Re-Id~\cite{zheng2016discriminatively,chen2016multi,geng2016deep} have two or three network branches, in order to obtain complementary properties. However, as our proposed structured graph Laplacian embedding enjoys the same requirement as the softmax loss and needs no extra recombination of the training data, thus the joint learning  only needs one network branch, which makes the CNN training very easy and more efficient.

\section{The Proposed Method}~\label{SectionProposedMethod}
In this section, we elaborate the structured graph Laplacian embedded  method for deep person Re-Id. First, we describe the proposed general framework. Then we revisit the conventional contrastive and triplet loss. After that, the proposed structured graph Laplacian algorithm will be presented, and followed by the optimization method.

\subsection{General Framework}
The problem of person Re-Id can be formulated as follows. Let $\{\vX_i,c_i\}_{i=1}^{N}$ be the set of input training data, where $\vX_i$ denotes the $i$-th raw input image data, $c_i \in \{1,2,\cdots, C\}$ denotes the corresponding person's identity label, $C$ is the number of identities, and $N$ is the number of training samples. The goal of training CNN is to learn filter weights and biases that minimize the identity error from the output layer, i.e., minimizing the distance between the same identities, and meanwhile maximizing the distance
between different identities. A recursive function for the $M$-th layer CNN model can be defined as follows:

\begin{equation}\label{ForwardParam}
\begin{split}
  &\vX_i^{(m)} = f(\vW^{(m)}*\vX_i^{(m-1)}) + \vb^{(m)} \\
  &i = 1,2,\ldots,N; m=1,2,\ldots,M; \vX_i^{(0)}=\vX_i,
\end{split}
\end{equation}
where $\vW^{(m)}$ denotes the filter weights of the $m$-th layer to be learned, $\vb^{(m)}$ refers to the corresponding biases, * denotes the convolution operation, $f(.)$ is an element-wise nonlinear activation function such as ReLU, and $\vX_i^{(m)}$ represents the feature map generated at layer $m$ for sample $\vX_i$. The total parameters of the CNN model can be denoted  as ${\mathcal W}=\{\vW^{(1)},\ldots,\vW^{(M)}; \vb^{(1)},\ldots,\vb^{(M)}\}$ for simplicity.

In this paper, we cast the person Re-Id task as a ranking problem.
We achieve this goal by embedding the structured graph Laplacian into the penultimate layer of the CNN model during the training process.
Embedding this graph Laplacian into the penultimate layer is equivalent to using the following loss function to train the CNN model:
\begin{equation}\label{jointcostfunction}
  \argmin_{\mathcal W} L=\sum_{i=1}^{N}\ell({\mathcal W},\vX_i,c_i) +
  \lambda{\mathcal R}({\mathcal X},\vc),
\end{equation}
where $\ell({\mathcal W},\vX_i,c_i)$ is the softmax loss for sample $\vX_i$,
and ${\mathcal R}({\mathcal X},\vc)$ denotes the structured graph Laplacian embedding.
The input to it includes ${\mathcal X}^{(k)}=\{\mathbf{x}_1, \cdots, \mathbf{x}_N\}$  which denotes the set of produced feature vectors at the penultimate layer for all the training samples, $\mathbf{x}_i$ denotes the feature vector of the input image $\mathbf{X}_i$, and $\vc=\{c_i\}_{i=1}^N$ is the set of corresponding labels.
Parameter $\lambda$ controls the trade-off between the softmax loss and the structured graph Laplacian embedding.

The softmax loss can encourage the deeply learned features to contain large inter-personal variations, while the proposed structured graph Laplacian embedding can make the learned features with more intra-personal compactness.


\subsection{Metric Embedding Revisit}
In this subsection, we  briefly review  the traditional \emph{contrastive loss} and \emph{triplet loss}.

\textbf{Contrastive Loss} is trained with the pairwise data $\{(\vx_i,\vx_j,\eta_{ij})\}$, $\eta_{ij} = 1$ if $\vx_i$ and $\vx_j$ are from the same class and  $\eta_{ij} = 0$ otherwise. Intuitively, the contrastive loss minimizes the distance between a pair of samples with the same class label and penalize the negative pair distances for being smaller than a predefined margin $\alpha$. Concretely, the loss function is defined as,
\begin{equation}\label{comparativeloss}
  L^{\nu} = \frac{1}{N}\sum_{i,j} \eta_{ij} D_{ij}^2 + (1-\eta_{ij})[\alpha - D_{ij}^2]_+,
\end{equation}
where $D_{ij}=||\vx_i - \vx_j||_2$, $N$ is the number of image pairs, $[x]_{+} = max\{x,0\}$.

\textbf{Triplet Loss} is trained with the triplet data $\{(\vx_i^{a},\vx_i^{p},\vx_i^{n})\}$, where $(\vx_i^{a},\vx_i^{p})$ share the same identity labels and $(\vx_i^{a},\vx_i^{n})$ have different identity labels. $\vx_i^{a}$ refers to as an anchor of a triplet. Intuitively, the training process encourages the network to make distance between $\vx_i^{a}$ and $\vx_i^{n}$  larger than that of  $\vx_i^{a}$ and $\vx_i^{p}$ plus the margin parameter $\tau$. The loss function is defined as,
\begin{equation}\label{tripletloss}
  L^{t} = \frac{1}{N}\sum_{i=1}^{N}[D_{ia,ip}^2 - D_{ia,in}^2 + \tau]_+,
\end{equation}
where $D_{ia,ip}=||\vx_{i}^{a} - \vx_{i}^{p}||_2$,
and $N$ is the number of sample triplets.


\subsection{Structured Graph Laplacian Embedding}~\label{laplacian}
As described in the previous subsection, we can clearly see that both the contrastive and triplet loss are constructed by different sample pair distances $D_{ij}$ in the training batch, which are just the special cases of the distance relationships among the training samples. Though they are very different, their basic elements are all the sample pair distances $D_{ij}$. This greatly motivates us to construct one complete graph to reformulate all these possible distance relationships in the training batches. By constructing a complete graph, we can utilize the global structured distance relationship information in the training batches, and thus the network can be optimized in the batch-global mode.

\begin{figure}[t]
 \centering
 \renewcommand{\arraystretch}{.9}
 \renewcommand{\tabcolsep}{.5mm}
 \includegraphics[width=7.5cm,height=3.5cm]{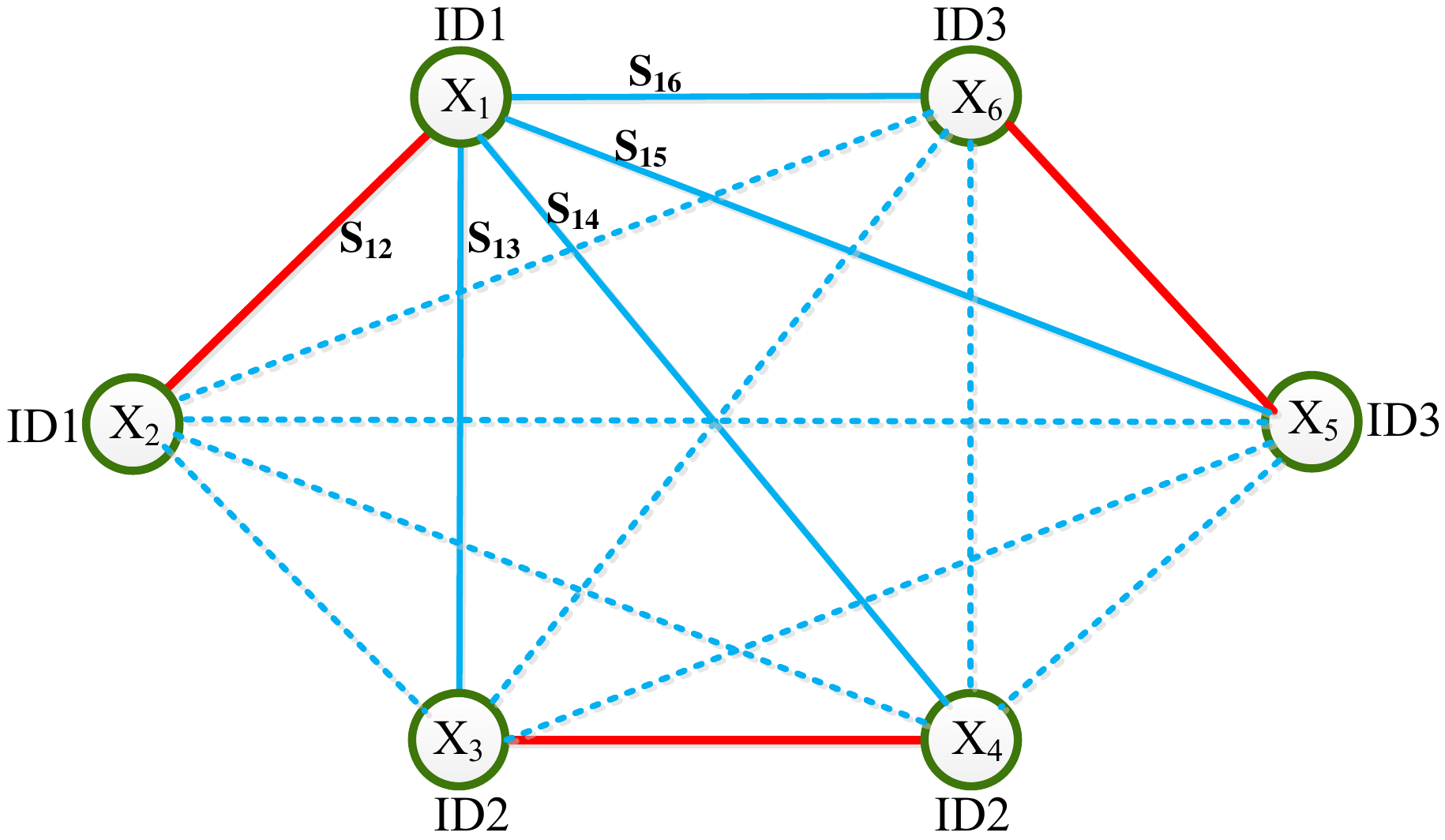}\\
 \caption{It illustrates the constructed complete graph with six nodes, where each node corresponds to one example in the training batch. The red edges and blue edges in the graph represent similar and dissimilar example pairs, respectively. The weight $S_{ij}$ for each distance edge is computed by the batch-global triple and contrastive loss.
 }
 \label{mainScope}
 \end{figure}

Therefore, we tend to propose the structured graph Laplacian embedding algorithm. As shown in Figure~\ref{mainScope}, we first construct one complete graph as follows,

\begin{equation}\label{unionLaplacianCost}
\begin{split}
  {\mathcal R}({\mathcal X},\vc) &=\sum_{i,j=1}^N S_{ij}||\vx_i-\vx_j||_2^2 \\
  &= 2\,tr(\vH\boldsymbol{\Psi}\vH^T),
\end{split}
\end{equation}
where ${\mathcal X}=\{\mathbf{x}_1, \cdots, \mathbf{x}_N\}$  denotes the set of produced feature vectors by the deep model for all the training samples, $\vc=\{c_i\}_{i=1}^N$ is the set of corresponding labels, and $N$ is the total number of training samples in the batch at here. $S_{ij}$ is the edge weight for the distance between node $i$ and $j$. For simplicity, we have rewrite it as the trace form in the second line.
$\vH=[\vx_1,\ldots,\vx_N]$, $\boldsymbol{\Psi} = \vG - (\vS+\vS^T)/2$, $\vG=diag(g_{11},\ldots,g_{NN})$, $g_{ii}=\sum_{j=1,j\neq i}^N \frac{S_{ij}+S_{ji}}{2},i=1,2,\ldots,N.$ $\boldsymbol{\Psi}$ is called the Laplacian matrix of $\vS$, and $tr(.)$ denotes the trace of a matrix.

As shown in Eq.~\eqref{unionLaplacianCost}, we construct a structured complete graph with $N$ nodes, where $N$ is the number of input training sample vectors in the batch. Then, we obtain $N \times N$ edges in the graph corresponding to $N \times N$ sample pair distances $||\vx_{i} - \vx_{i}||_2^2$. Note that the edge between oneself is discarded in Figure~\ref{mainScope}, and we always set $S_{ii}=0$. Based on the constructed complete graph, we define the new \emph{ batch-global contrastive loss} and \emph{batch-global triplet loss} in the training batches. This step enables the proposed method to take full advantages of the distance relationships in the whole batch, where the relationships can be reflected in the weight matrix $\vS$ according to different requirements.



\textbf{First}, we introduce the \textbf{batch-global contrastive loss} in each training batch as follows,
\begin{equation}\label{globalcontrastive}
\begin{split}
  {\mathcal R}^\nu &= \sum_{i,j=1; \atop i\neq j}^{N}(\eta_{ij}D_{ij}^2 + (1-\eta_{ij})[\alpha - D_{ij}^2]_+) \\
  &\doteq\sum_{i,j=1}^N S_{ij}^\nu D_{ij}^2
  = \sum_{i,j=1}^N S_{ij}^\nu||\vx_{i} - \vx_{i}||_2^2,
\end{split}
\end{equation}
where the weight matrix $S_{ij}^\nu$ in Eq.~\eqref{globalcontrastive} for the \emph{batch-global contrastive loss} can be deduced  as follows,
\begin{equation}\label{comparativeSij}
  S_{ij}^\nu=\left\{
  \begin{aligned}
    &1,\quad c_i=c_j,i\neq j\\
    &-\delta[\alpha - D_{ij}^2], \quad c_i\neq c_j, i\neq j,\\
    &0, \quad i=j.
  \end{aligned}
\right.
\end{equation}
$D_{ij}=||\vx_{i} - \vx_{i}||_2$, and $\delta[.]$ is an indicator function which takes one if the argument is bigger than zero, and zeros otherwise.

\textbf{Second}, we introduce the \textbf{batch-global triplet loss} in the training batches as Eq.~\eqref{globaltriplet},
\begin{equation}\label{globaltriplet}
\begin{split}
  {\mathcal R}^t &= \sum\limits_{i,j,k=1; \atop c_i=c_j \neq c_k}^{N}[D_{ij}^2 - D_{ik}^2 + \tau]_+ \\
  &\doteq \sum_{i,j=1}^N S_{ij}^tD_{ij}^2
  =\sum_{i,j=1}^N S_{ij}^t||\vx_{i} - \vx_{i}||_2^2.
\end{split}
\end{equation}
Similarly, the weight matrix $S_{ij}^t$  can be deduced  as:
\begin{equation}\label{tripletSij}
  S_{ij}^t=\left\{
  \begin{aligned}
    &\sum_{k=1,\atop c_i=c_j \neq c_k}^N \delta[D_{ij}^2 - D_{ik}^2 + \tau],  i\neq j,\\
    &-\sum_{k=1,\atop c_i=c_k\neq c_j.}^N \delta[D_{ik}^2 - D_{ij}^2 + \tau],  i\neq j,\\
    &0, \quad i=j.
  \end{aligned}
\right.
\end{equation}

\emph{Note that we have ignored one constant when deducing from the first line to second line in Eq.~\eqref{globalcontrastive} and Eq.~\eqref{globaltriplet} because the constant has no effect on the optimization. We use the symbol ``$\doteq$'' instead of ``$=$'' for differentiation..}

In order to further explore their structured distance relationships, we have combined the \emph{batch-global contrastive loss} and \emph{batch-global triplet loss} to form a new structured graph Laplacian loss as follows,

\begin{equation}\label{globalcomparison}
\begin{split}
   {\mathcal R}({\mathcal X},\vc) &= {\mathcal R}^t+ \beta{\mathcal R}^\nu \\
  &=\sum_{i,j=1}^N (S_{ij}^t + \beta S_{ij}^\nu)||\vx_{i} - \vx_{i}||_2^2\\
  & =\sum_{i,j=1}^N S_{ij} ||\vx_{i} - \vx_{i}||_2^2,
\end{split}
\end{equation}
where $\beta$ is one parameter to balance the batch-global contrastive and triplet loss, and $S_{ij}=S_{ij}^t + \beta S_{ij}^\nu$. Notice that the $L2$ normalization is conducted in each row when the weight matrix $S_{ij}^\nu$, $S_{ij}^t$ are constructed.

\textbf{Discussion:}
By constructing one complete graph to form the traditional contrastive and triplet loss into the structured graph Laplacian form as illustrated in Eq.~\eqref{globalcontrastive}, Eq.~\eqref{globaltriplet} and Eq.~\eqref{globalcomparison},
our proposed algorithm has the following merits:
1) Since we have constructed one complete graph in the training batch, the proposed algorithm can take full advantages of the samples' distance relationships in the training batch, and thus can be optimized in the batch-global mode to obtain more discriminative features.
2) As shown in Figure~\ref{mainScope}, we construct a complete graph with six nodes corresponding to six input training sample vectors of three different identities, our methods can utilized 24 sample triplets in total at once, while traditional triplet loss needs 24 times' input for three different network branches to use the same amount of structured distance information. Thus, we utilize much more distance structure information in each batch, which results in faster convergence speed, theoretically. 3) Our algorithm uses only one network branch, and no extra complex data expansion and redundant memory is needed. Thus, it is easier to be trained and implemented compared with the traditional siamese and triplet networks.

\subsection{Optimization}

We use the back-propagation method to train the CNN model, which is carried out in one mini-batch. Therefore, we need to calculate the gradients of the loss function with respect to the features of the corresponding layers.




The gradient of ${\mathcal R}({\mathcal X},\vc)$ in Eq.~\eqref{unionLaplacianCost} with respect to ${\mathcal W}$ is:
\begin{equation}\label{gradient_all}
  \frac{\partial {\mathcal R}}{\partial {\mathcal W}}= \sum_{i=1}^N \frac{\partial {\mathcal R}}{\partial \vx_{i}}\cdot \frac{\partial \vx_{i}}{\partial {\mathcal W}},
\end{equation}
where
\begin{equation}\label{gradient_xi}
  \frac{\partial {\mathcal R}}{\partial \vx_{i}}=2\vH(\boldsymbol{\Psi} + \boldsymbol{\Psi}^T)_{(:,i)}=4\vH \boldsymbol{\Psi}_{(:,i)},
\end{equation}
and $\boldsymbol{\Psi}_{(:,i)}$ denotes the $i$-th column of matrix $\boldsymbol{\Psi}$. The definition of matrix $\Psi$ and $\vH$ is the same as in Section~\ref{laplacian} in Eq.~\eqref{unionLaplacianCost}.  More details about the optimization for the proposed structured graph Laplacian embedding can refer to~\cite{niyogi2004locality}.

The total gradient of Eq.~\eqref{jointcostfunction} with respect to ${\mathcal W}$ is simply the combination of the gradient from the conventional softmax loss and the above gradient from the structured graph Laplacian embedding in Eq.~\eqref{globalcomparison}. While in Algorithm~\ref{Trainingalgorithm}, we just illustrate the training algorithm for the structured graph Laplacian embedding method, since softmax loss is an well developed function in ``Caffe'' package~\cite{jia2014caffe}.

%
%

\begin{algorithm}
\caption{The training algorithm for the structured graph Laplacian embedding method.}
\label{Trainingalgorithm}
\KwIn{Training data~$\{ \vX_i, c_i \}_1^N$. Initialized parameters ${\mathcal W}$  for the CNN network. Parameter $\lambda$, $\alpha$, $\tau$, $\beta$ and learning rate $\mu^t$. The number of iteration $t\leftarrow 0$.}
\KwOut{The network parameters ${\mathcal W}$.}

\For{$t = 1,2, \ldots, T$}
{
Calculate the weight matrix $S^t_{ij}$ for the global triplet cost according to Eq.~\eqref{tripletSij}\;

Calculate the weight matrix $S^\nu_{ij}$ for the global contrastive cost  according to Eq.~\eqref{comparativeSij}\;

Calculate the weight matrix for our proposed model in Eq.~\eqref{globalcomparison}, $S_{ij}=S^t_{ij}+\beta S^\nu_{ij}$\;

Calculate the Laplacian matrix \scriptsize{$\boldsymbol{\Psi} = \vG - (\vS+\vS^T)/2$}\;

\normalsize{\textbf{Forward Propagation:}}

\normalsize{Calculate the structured graph Laplacian embedding ${\mathcal R}$ by Eq.~\eqref{unionLaplacianCost}}\;

\textbf{Backward Propagation:}

{
Calculate the backpropagation gradient $\frac{\partial {\mathcal R}}{\partial {\mathcal W}}$  based on Eq.~\eqref{gradient_all} and Eq.~\eqref{gradient_xi}.\;

Update the parameters ${\mathcal W}^t = {\mathcal W}^{t-1}-\mu_t \cdot \frac{\partial {\mathcal R}}{\partial {\mathcal W}}$.\;
}

}
\end{algorithm}

\section{Experiments}~\label{SectionExperiments}
We conduct experiments on four popular person Re-Id datasets with three base network architectures, namely AlexNet, ResNet50 and the DGDNet. Experimental results and the detailed analysis of the proposed ranking Laplacian objectives are all illustrated in this section.

\subsection{Datasets and protocols}~\label{datasetsandproto}
In this section, we use four widely used person Re-Id benchmark datasets to evaluate our proposed methods, including two relatively small datasets, namely 3DPES~\cite{baltieri20113dpes} and CUHK01~\cite{li2013locally}, and two large datasets of CUHK03~\cite{li2014deepreid} and Market-1501~\cite{zheng2015scalable}. All the datasets contain a set of persons, each of whom has several images captured by different cameras. In the following, we will give brief descriptions of these four datasets:

\begin{description}
 \item{\textbf{\textit{3DPES dataset}}} \cite{baltieri20113dpes} includes 1,011 images of 192 persons captured from 8 outdoor cameras with significantly different viewpoints. The number of images for each person varies from 2 to 26.
     This is one of the most challenging datasets for the person re-id task due to its huge variance and discrepancy.

 \item{\textbf{\textit{CUHK01 dataset}}} \cite{li2013locally} contains 971 persons from two camera views in a campus environment. Camera view A captures frontal or back views of a person while camera B captures the person's profile views. Each person has four images with two from each camera.

 \item{\textbf{\textit{CUHK03 dataset}}} \cite{li2014deepreid} is one of the largest person Re-Id benchmark datasets recently. It contains 13,164 images of 1,360 identities, and the images were captured by five different pairs of camera views in the campus.

 \item{\textbf{\textit{Market-1501 dataset}}} \cite{zheng2015scalable} contains 1,501 individuals, including 19,732 gallery images and 12,936 training images captured by 6 cameras, the persons in each image are produced by the DPM detector~\cite{felzenszwalb2010object}.
\end{description}


In the testing phase, the score between each probe image and the gallery image is computed just by the euclidean distance. We adopt the widely used cumulative match curve(CMC) metric for quantitative evaluations.  For datasets 3DPES, CUHK01 and CUHK03, we have used the same settings as~\cite{xiao2016learning} to do experiments and evaluate their results. While for the Market-1501 dataset, we use the toolbox provided by \cite{zheng2015scalable}, and the mean Average Precision(mAP) is also computed. We also randomly drew roughly $20\%$ of all these images for validation. Notice that both the training and validation identities have no overlap with the test ones. We adopt the single-shot mode and evaluate all the datasets
under 20 random train/test splits.

\subsection{Experimental Setup}
In the experiments, the parameters $\tau$, and  $\alpha$ are both set to be $1.0$, empirically. The trade-off parameters $\lambda$ and $\beta$ are set to be $0.6$ and 0.1, respectively. Our code is implemented with the ``Caffe'' package~\cite{jia2014caffe}, and all the experiments are done on the GPU TitanX with 12GB memory. To make the training more efficient, every two images of the same person from two different camera views are bundled up, and then shuffle them in the datasets. The AlexNet and ResNet50 in the experiments are trained by fine-tuning from the publicly released model trained on the ImageNet Datasets, while the DGDNet is trained from scratch. On datasets CUHK03, CUHK01 and 3DPES, we use the same setting as~\cite{xiao2016learning} for fair comparison, which first jointly train the model on all the datasets, and then finetune on each of them. While for Market-1501 dataset, all experimental results are trained from this single datasets.

\subsection{Comparison with state-of-the-art methods}~\label{SectionResultComparison}
We compare the results of our proposed algorithm with these state-of-the-art ones on four datasets. Table~\ref{3DPESdataset},~\ref{CUHK01dataset},~\ref{CUHK03dataset}, and \ref{Market-1501dataset} show the results on 3DPES, CUHK01, CUHK03 and Market-1501 datasets, respectively, using the rank 1, 5, 10, 20 accuracies. Each table includes the most recently reported evaluation results, including some traditional metric learning methods~\cite{mignon2012pcca,xiong2014person,chen2016similarity,zhang2016learning,liao2015person}, and much more deep learning based methods~\cite{xiao2016learning,wang2016joint,wu2016deep,varior2016gated,xiao2016cross,zheng2016discriminatively,li2014deepreid,ahmed2015improved,varior2016siamese,zheng2017pose,barbosa2017looking}.
Due to the space limited, only the most representative and competitive ones are chosen to be compared with. Since we have used three base networks in the experiments, then we define them as ``ours-AlexNet'', ``ours-ReNet50'' and ``ours-DGDNet'', respectively.   From the experimental results in Table~\ref{3DPESdataset},~\ref{CUHK01dataset},~\ref{CUHK03dataset}, and \ref{Market-1501dataset}, we can obtain the following observations: 1) Our proposed joint learning method outperforms all the compared methods, and we achieve state-of-the-art performances on the four benchmark datasets. 2) Compared within the three base networks, we can clearly see that the joint learning methods based on the relatively small DGDNet can achieve the best performance, and the ours-AlexNet gets the worst performance among them. 3) When dividing all the compared methods into two categories, we can find that, the traditional metric learning methods with the handcraft features can achieve relatively comparable or slightly worse performances with the deep learning based methods on the small datasets, such as 3DPES and CUHK01, while on the larger datsets (CUHK03 and Market-1501), the deep learning based methods can outperform the traditional metric learning based methods over a large margin.


\begin{table}[t]
  \setlength{\tabcolsep}{8pt}
  \centering
  \caption{\small{Experimental results on 3DPES dataset.}} \label{3DPESdataset}
  \begin{tabular}{p{2.8cm}p{0.6cm}p{0.6cm}p{0.7cm}p{0.7cm}}\\
		\toprule
  \multirow{2}{*}{Method} &
  \multicolumn{4}{c}{3DPES (p=92)} \\
  \cline{2-5}
  &\footnotesize{rank1} &\footnotesize{rank5} &\footnotesize{rank10}  &\footnotesize{rank20}\\
  \midrule
  LSMLEC\cite{kostinger2012large} &$34.2$ &$58.7$ &$69.6$ &$80.2$ \\
  PCCA \cite{mignon2012pcca} &$43.5$ &$71.6$ &$81.8$ &$91.0$  \\
  LFDA\cite{pedagadi2013local} &$45.5$ &$69.2$ &$70.1$ &$82.1$   \\
  KernelM\cite{xiong2014person} &$54.0$ &$77.7$ &$85.9$ &$92.4$ \\
  Ensembles\small{\cite{paisitkriangkrai2015learning}} &$53.3$ &$76.8$ &$85.7$ &$91.4$  \\
  DGDNet\cite{xiao2016learning} &$55.2$ &$76.4$ &$84.9$ &$91.9$ \\
  SCSP\cite{chen2016similarity} &$57.3$ &$78.6$ &$86.5$ &$93.6$ \\

  \midrule
  ours-AlexNet &$42.0$ &$64.6$ &$74.1$ &$84.3$ \\
  ours-ResNet50 &$52.2$ &$79.5$ &$\textbf{88.1}$ &$\textbf{93.9}$ \\
  \textbf{ours-DGDNet} &$\textbf{61.0}$ &$\textbf{80.3}$ &$87.4$ &$93.1$ \\
  \bottomrule
  \end{tabular}
\end{table}

\begin{table}[t]
  \setlength{\tabcolsep}{8pt}
  \centering
  \caption{\small{Experimental results on CUHK01 dataset.}} \label{CUHK01dataset}
  \begin{tabular}{p{2.8cm}p{0.6cm}p{0.6cm}p{0.7cm}p{0.7cm}}\\
		\toprule
  \multirow{2}{*}{Method} &
  \multicolumn{4}{c}{CUHK01 (p=486)} \\
  \cline{2-5}
   &\footnotesize{rank1} &\footnotesize{rank5} &\footnotesize{rank10}  &\footnotesize{rank20}\\
  \midrule
  KHPCA\cite{prates2016kernel} &$38.3$ &$66.8$ &$77.7$ &$86.8$ \\
  MIRROR\cite{chen2015mirror} &$40.4$ &$64.6$ &$75.3$ &$84.1$ \\
  Ahmed\cite{ahmed2015improved} &$47.5$ &$71.6$ &$80.3$ &$87.5$\\
  KernelM\cite{xiong2014person} &$49.6$ &$74.7$ &$83.8$ &$91.2$  \\
  Ensembles\cite{paisitkriangkrai2015learning} &$53.4$ &$76.4$ &$84.4$ &$90.5$ \\
  SCSP\cite{chen2016similarity} &$56.8$ &$87.6$ &$89.5$ &$92.3$\\
  HGauD\cite{matsukawa2016hierarchical}  &$57.8$ &$79.1$ &$86.2$ &$92.1$  \\
  JointSC\cite{wang2016joint} &$52.2$ &$83.7$ &$89.5$ &$94.3$ \\
  LOMO\cite{liao2015person} &$63.2$ &$83.9$ &$90.0$ &$94.2$  \\
  DGDNet\cite{xiao2016learning} &$66.6$ &$85.5$ &$90.5$ &$94.2$ \\
  NullSpace\cite{zhang2016learning} &$69.1$ &$86.9$ &$91.8$ &$95.4$ \\
  \midrule
  ours-AlexNet &$44.0$ &$67.4$ &$75.8$ &$83.8$ \\
  ours-ResNet50 &$58.2$ &$79.8$ &$86.7$ &$92.4$ \\
  \textbf{ours-DGDNet} &$\textbf{70.9}$ &$\textbf{89.8}$ &$\textbf{93.5}$ &$\textbf{95.9}$\\
  \bottomrule
  \end{tabular}
\end{table}

\begin{table}[t]
  \setlength{\tabcolsep}{8pt}
  \centering
  \caption{\small{Experimental results on CUHK03 labeled dataset.}} \label{CUHK03dataset}
  \vspace{-.8em}
  \begin{tabular}{p{2.8cm}p{0.6cm}p{0.6cm}p{0.7cm}p{0.7cm}}\\
		\toprule
 \multirow{2}{*}{Method} &
  \multicolumn{4}{c}{CUHK03 (p=100)} \\
  \cline{2-5}
  &\footnotesize{rank1} &\footnotesize{rank5} &\footnotesize{rank10}  &\footnotesize{rank20}\\
  \midrule
  UnsupSal\cite{zhao2013unsupervised} &$8.8$ &$24.1$ &$38.3$ &$53.4$  \\
  LSMLEC\cite{kostinger2012large} &$14.2$ &$48.5$ &$52.6$ &$--$  \\
  Deepreid\cite{li2014deepreid} &$20.6$ &$51.5$ &$66.5$ &$80.0$  \\
  LOMO\cite{liao2015person} &$52.2$ &$82.2$ &$92.1$ &$96.2$  \\
  KernelM\cite{xiong2014person} &$48.2$ &$59.3$ &$66.4$ &$--$ \\
  JointSC\cite{wang2016joint} &$52.2$ &$83.7$ &$89.5$ &$94.3$ \\
  Ahmed\cite{ahmed2015improved} &$54.7$ &$86.5$ &$94.0$ &$96.1$ \\
  LSTMSia\cite{varior2016siamese} &$57.3$ &$80.1$ &$88.3$ &$--$  \\
  Ensembles\small{\cite{paisitkriangkrai2015learning}} &$62.1$ &$89.1$ &$94.3$ &$97.8$\\
  NullSpace\cite{zhang2016learning} &$58.9$ &$85.6$ &$92.5$ &$96.3$ \\
  DeepLDAF\cite{wu2016deep} &$63.2$ &$90.0$ &$92.7$ &$97.6$ \\
  GatedSia\cite{varior2016gated} &$68.1$ &$88.1$　&$94.6$ &$--$\\
  CDKT \cite{xiao2016cross} &$74.8$ &$94.7$　&$97.8$ &$98.9$\\
  DGDNet\cite{xiao2016learning} &$78.3$ &$94.6$ &$97.6$ &$99.0$ \\
  DisEmbed\cite{zheng2016discriminatively} &$83.4$ &$97.1$　&$98.1$ &$--$\\
  \midrule
  ours-AlexNet &$56.2$ &$85.3$ &$93.3$ &$96.9$ \\
  ours-ResNet50 &$73.2$ &$93.7$ &$97.2$ &$98.7$ \\
  \textbf{ours-DGDNet} &$\textbf{84.7}$ &$\textbf{97.4}$ &$\textbf{98.9}$ &$\textbf{99.5}$\\
  \bottomrule
  \end{tabular}
\end{table}

\begin{table}[t]
  \setlength{\tabcolsep}{5pt}
  \centering
  \caption{\small{Experimental results on Market-1501 dataset.}} \label{Market-1501dataset}
  \begin{tabular}{p{2.8cm}p{0.6cm}p{0.6cm}p{0.7cm}p{0.7cm}p{0.7cm}}\\
		\toprule
  \multirow{2}{*}{Method} &
  \multicolumn{5}{c}{Market-1501 (p=751)} \\
  \cline{2-6}
   &\footnotesize{rank1} &\footnotesize{rank5} &\footnotesize{rank10}  &\footnotesize{rank20} &mAP\\
  \midrule
  BoWKis\cite{zheng2015scalable} &$44.4$ &$63.9$ &$72.2$ &$78.9$ &$20.76$ \\
  WARCR\cite{jose2016scalable} &$45.2$ &$68.1$ &$76.0$ &$84.0$ &$--$\\
  TempAda\cite{martinel2016temporal} &$47.9$ &$--$ &$--$ &$--$ &$22.31$\\
  SCSP\cite{chen2016similarity} &$51.9$ &$--$ &$--$ &$--$ &$26.35$  \\
  NullSpace\cite{zhang2016learning} &$55.43$ &$--$ &$--$ &$--$ &$29.87$ \\
  LSTMSia\cite{varior2016siamese} &$61.6$ &$--$ &$--$ &$--$ &$35.30$\\
  GatedSia\cite{varior2016gated}  &$65.9$ &$--$ &$--$ &$--$  &$39.55$\\
  SynDeep\cite{barbosa2017looking} &$73.9$ &$88.0$　&$92.2$ &$96.2$ &$47.89$\\
  PIE-Kis\cite{zheng2017pose} &$79.3$ &$90.8$ &$94.4$ &$96.5$ &$55.95$ \\
  DisEmbed\cite{zheng2016discriminatively} &$79.5$ &$90.9$ &$94.1$ &$96.2$ &$59.87$ \\
  \midrule
  ours-AlexNet  &$71.1$ &$86.0$ &$90.1$ &$93.1$ &$44.14$\\
  ours-ResNet50 &$72.3$ &$86.4$ &$90.6$ &$94.2$ &$46.78$\\
  \textbf{ours-DGDNet} &$\textbf{83.6}$ &$\textbf{93.7}$ &$\textbf{96.1}$ &$\textbf{97.8}$ &$\textbf{63.34}$\\
  \bottomrule
  \end{tabular}
\end{table}

\begin{table}[t]
  \setlength{\tabcolsep}{5pt}
  \centering
  \caption{\small{Parameter \# of each network architecture.}} \label{Parameter}
  \begin{tabular}{p{2.3cm}p{1.5cm}p{1.5cm}p{1.5cm}}\\
		\toprule
  Network &AlexNet &ResNet50 &DGDNet\\
  \midrule
  Parameter \#  &55.57M &$23.84$M &$5.84$M \\
  \bottomrule
  \end{tabular}
\end{table}

Given the above observations, we can see that the ResNet50 can not work as well as the DGDNet on the Re-Id datasets, which is different from the observations in other visual recognition tasks. This is reveals thet the datasets, even with the recent large CUHK03 and Market-1501 datasets, are still relatively very small to make the deep models release their full potential. The deep network with too many parameters is easy to overfit the datasets. As illustrated in Table~\ref{Parameter}, we compared the number of parameters for the used three base network architectures. We can also see that the network with the relatively smallest size of parameter quantity has the best performance, without considering the difference in the network architectures. To further verify the effect of parameter quantity on the performances, we did experiments with the same AlexNet on the large Market-1501 datasets, but with different parameter quantity as  illustrated in Table~\ref{AlexNet-Market-1501dataset}. We have set the number of output in the last two fully connected layers as 4096, 2048, 1024 and 512, then the parameter quantity drops dramatically from 55.57M to 8.32M, and we got the best performance by the AlexNet with 1024 fully connected outputs.

\begin{table*}
 \setlength{\tabcolsep}{6pt}
\centering
\caption{Experimental results with different network architectures and loss function on four person Re-Id benchmark datasets.}\label{MulitDatasets}
\begin{tabular}{p{3.3cm}|p{0.6cm}|p{0.5cm}p{0.5cm}p{0.6cm}|p{0.5cm}p{0.5cm}p{0.6cm}|p{0.5cm}p{0.5cm}p{0.6cm}|p{0.5cm}p{0.6cm}p{0.6cm}p{0.6cm}}
\hline
\multirow{2}{3.3cm}{Methods} &\multirow{2}{0.6cm}{dim} &\multicolumn{3}{|c|}{CUHK03} & \multicolumn{3}{|c|}{3DPES} & \multicolumn{3}{|c|}{CUHK01}   & \multicolumn{4}{|c}{Market-1501}\\
\cline{3-15}

&\quad &\footnotesize{rank1} &\footnotesize{rank5} &\footnotesize{rank10}    &\footnotesize{rank1} &\footnotesize{rank5} &\footnotesize{rank10}   &\footnotesize{rank1} &\footnotesize{rank5} &\footnotesize{rank10}  &\footnotesize{rank1} &\footnotesize{rank5} &\footnotesize{rank10}   &\footnotesize{Map}\\
\hline
BGCL(AlexNet, fc7)   &$4096$  &$4.7$ &$17.2$ &$23.7$ &$7.7$ &$24.5$ &$34.7$  &$2.0$ &$12.2$ &$23.8$ &$6.8$ &$14.3$ &$23.3$  &$2.97$\\
BGTL(AlexNet, fc7)    &$4096$ &$11.4$ &$36.5$ &$52.0$ &$13.3$ &$33.5$ &$47.1$  &$11.9$ &$29.6$ &$40.9$ &$8.8$ &$24.9$ &$35.6$  &$4.01$\\
Softmax(AlexNet, fc7)  &$4096$   &$ {34.1}$ &$ {65.4}$ &$ {78.6}$ &$ {35.7}$ &$ {58.3}$ &$ {67.7}$  &$ {26.6}$ &$ {49.6}$ &$ {59.4}$ &$ {50.6}$ &$ {71.3}$ &$ {79.2}$  &$ {26.53}$\\

ours(AlexNet, fc7)  &$4096$  &$\textbf{56.2}$ &$\textbf{85.3}$ &$\textbf{93.3}$ &$\textbf{42.0}$ &$\textbf{64.6}$ &$\textbf{74.1}$  &$\textbf{44.0}$ &$\textbf{67.4}$ &$\textbf{75.8}$  &$\textbf{71.1}$ &$\textbf{86.0}$ &$\textbf{90.1}$  &$\textbf{44.14}$\\
\hline
\hline
BGCL(ResNet50, p5) &$2048$ &$7.8$ &$25.2$ &$39.4$ &$20.3$ &$39.7$ &$52.2$  &$7.6$ &$17.8$ &$24.9$ &$16.8$ &$35.9$ &$47.1$  &$7.58$\\
BGTL(ResNet50, p5)  &$2048$   &$19.2$ &$48.7$  &$64.2$ &$25.0$ &$51.3$ &$63.9$ &$20.5$ &$45.3$ &$58.2$ &$19.8$ &$39.5$ &$50.2$  &$9.21$\\
Softmax(ResNet50,p5) &$2048$   &$ {46.1}$ &$ {76.6}$ &$ {86.6}$ &$ {50.1}$ &$ {74.2}$ &$ {83.4}$  &$ {41.3}$ &$ {63.9}$ &$ {72.2}$ &$ {68.8}$ &$ {83.9}$ &$ {88.1}$  &$ {40.73}$\\

ours(ResNet50, p5) &$2048$   &$\textbf{73.2}$ &$\textbf{93.7}$ &$\textbf{97.2}$ &$\textbf{52.2}$ &$\textbf{79.5}$ &$\textbf{88.1}$  &$\textbf{58.2}$ &$\textbf{79.8}$ &$\textbf{86.7}$
&$\textbf{72.3}$ &$\textbf{86.4}$ &$\textbf{90.6}$  &$\textbf{46.78}$\\
\hline
\hline
BGCL(DGD, fc7)  &$256$  &$13.2$ &$39.2$ &$57.9$ &$4.1$ &$11.7$ &$17.5$  &$13.8$ &$26.9$ &$32.7$ &$8.6$ &$21.0$ &$28.7$  &$4.61$\\
BGTL(DGD, fc7) &$256$  &$54.3$ &$87.0$ &$93.4$ &$31.4$ &$56.7$ &$69.7$  &$35.8$ &$69.8$ &$79.4$ &$34.3$ &$56.5$ &$66.4$  &$17.56$\\
BGCTL(DGD, fc7) &$256$  &$56.8$ &$87.9$ &$94.4$ &$33.1$ &$58.7$ &$69.8$  &$36.7$ &$69.6$ &$80.2$ &$36.1$ &$58.0$ &$68.4$  &$19.14$\\
Softmax(DGD, fc7) &$256$   &$ {82.4}$ &$ {95.9}$ &$ {98.0}$ &$ {58.3}$ &$ {79.2}$ &$ {86.7}$  &$ {68.9}$ &$ {87.8}$ &$ {92.4}$ &$ {81.7}$ &$ {92.5}$ &$ {95.1}$  &$ {61.31}$\\

ours(DGD, fc7)  &$256$  &$\textbf{84.7}$ &$\textbf{97.4}$ &$\textbf{98.9}$ &$\textbf{61.0}$ &$\textbf{80.3}$ &$\textbf{87.4}$  &$\textbf{70.9}$ &$\textbf{89.8}$ &$\textbf{93.5}$ &$\textbf{83.6}$ &$\textbf{93.7}$ &$\textbf{96.1}$  &$\textbf{63.34}$\\
\hline
\end{tabular}
\end{table*}

\begin{table}[t]
  \setlength{\tabcolsep}{5pt}
  \centering
  \caption{\small{Experimental results on Market-1501 dataset.}} \label{AlexNet-Market-1501dataset}
  \begin{tabular}{p{2.2cm}p{1.2cm}|p{0.7cm}p{0.7cm}p{0.8cm}p{0.6cm}}\\
		\toprule
  \multirow{2}{*}{Method} &
  \multicolumn{5}{c}{Market-1501 (p=751)} \\
  \cline{2-6}
   &\footnotesize{\#Param} &\footnotesize{rank1} &\footnotesize{rank5} &\footnotesize{rank10}  &mAP\\
  \midrule
  AlexNet-4096 &$55.57$M &$50.6$　&$71.3$ &$79.2$ &$26.53$\\
   \midrule
  AlexNet-2048 &$25.57$M &$58.8$　&$79.1$ &$85.0$ &$34.42$\\
   \midrule
  AlexNet-1024 &$13.57$M &$\textbf{60.4}$　&$\textbf{80.4}$ &$\textbf{86.5}$ &$\textbf{35.69}$\\
   \midrule
  AlexNet-512 &$8.32$M &$59.7$　&$79.5$ &$85.9$ &$35.50$\\
  \bottomrule
  \end{tabular}
\end{table}


\subsection{Analysis of the proposed method}~\label{SectionAnalysis}
To validate the effectiveness of the joint learning of the  proposed structured graph Laplacian embedding with the softmax loss, we have used the proposed batch-global contrastive loss (denoted as BGCL), batch-global triplet loss (denoted as BGTL), their combination as defined in Eq.~\eqref{globalcomparison}(denoted as BGCTL), and softmax loss(denoted as Softmax) as the supervision separately, to train the deep CNN model for person Re-Id, with different base network. Then, the joint supervision of the softmax loss and the proposed structured graph Laplacian embedding in Eq.~\eqref{globalcomparison} is used. The experimental results are illustrated in Table~\ref{MulitDatasets} to reveal how much performances are obtained by each method with different base network architectures.

As can be seen from Table~\ref{MulitDatasets}, we can get the following observations: 1) The joint learning method with all base networks can obtain better performance than each separate one, on all the datasets. This shows  that the proposed structured graph Laplacian embedding can help further improve the  CNN training. 2) When we use the AlexNet and the ResNet50, the improvement of the joint learning from  using the only softmax loss is more significant, while with the DGDNet, it improves relatively less. This may be the following reasons: the parameter quantities of AlexNet and ResNet50 are so large and easy to overfit the datasets, while integrating the graph Laplacian as another supervision could reduce the heavy overfitting.
3)The method BGTL can work better than BGCL on these person Re-Id datasets based on the three deep CNN model, and when combined them(BGCTL), about 2\% improvements  will be obtained with the DGDNet.

 \begin{figure}[t]
 \centering
 \renewcommand{\arraystretch}{.9}
 \renewcommand{\tabcolsep}{.5mm}
\begin{tabular}{cc}
 \includegraphics[width=4cm,height=3cm]{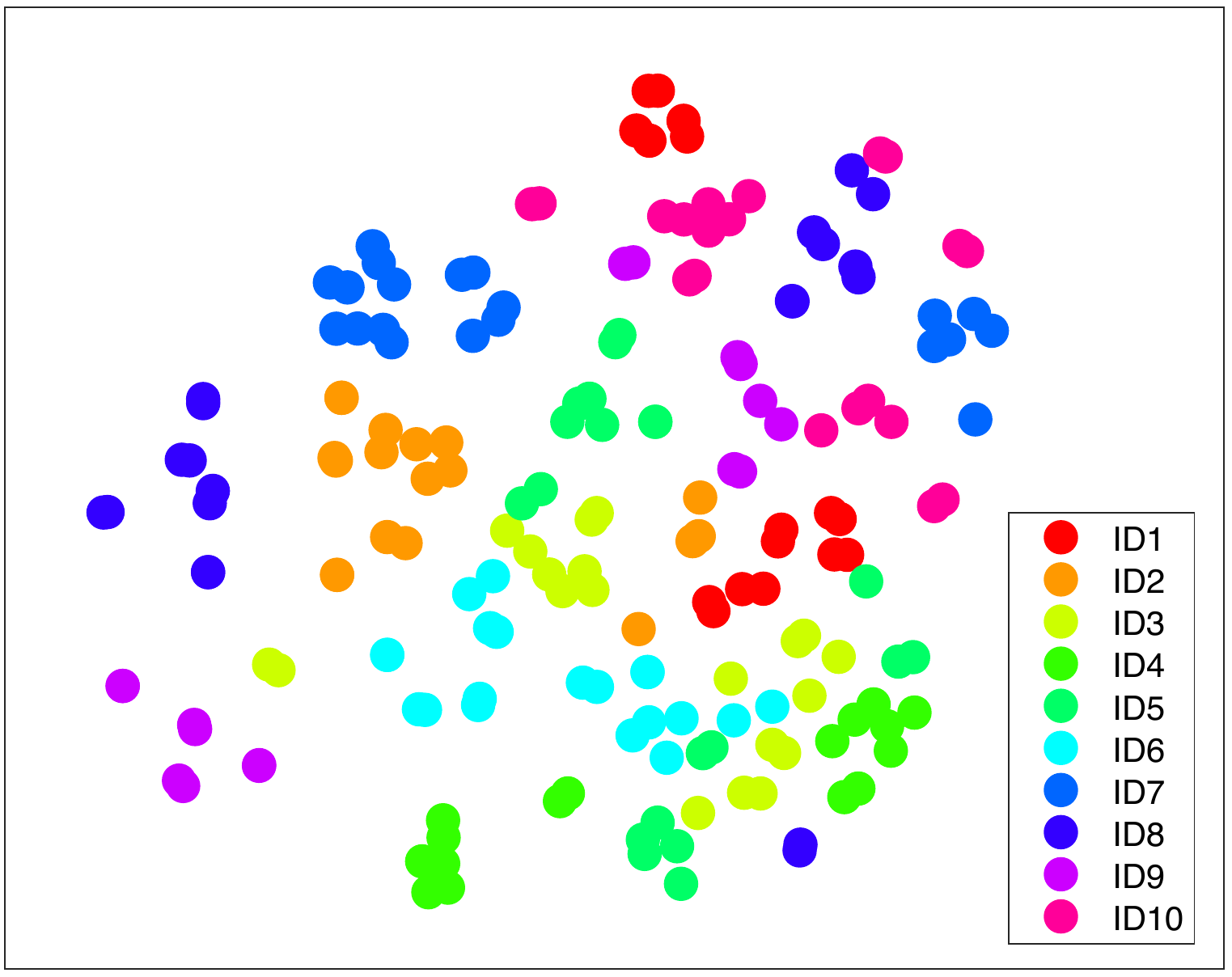}&
 \includegraphics[width=4cm,height=3cm]{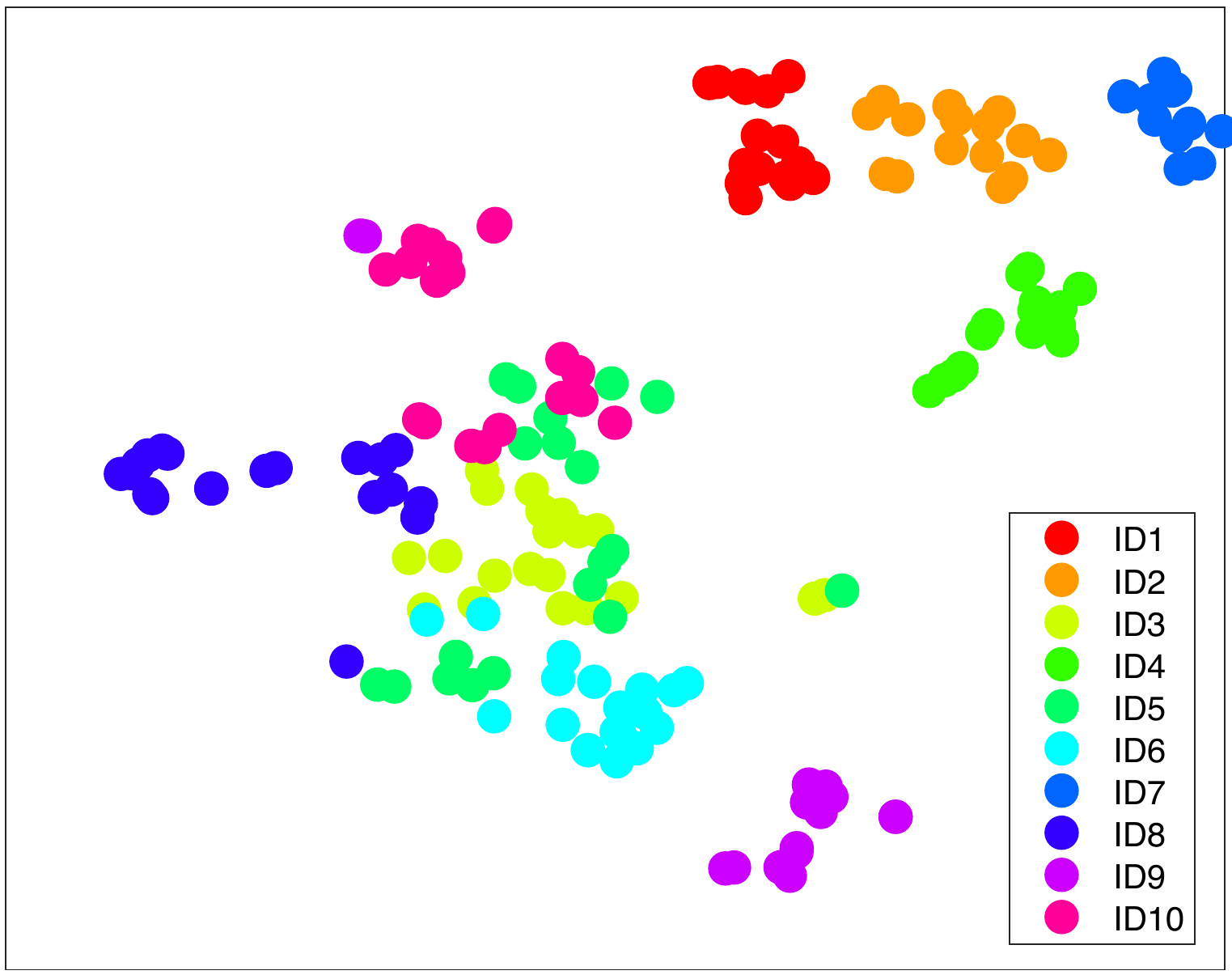}\\
  \footnotesize{(a)BGCL(ResNet50,2048-D)}. & \footnotesize{(b) BGTL(ResNet50,2048-D)} \\
 \includegraphics[width=4cm,height=3cm]{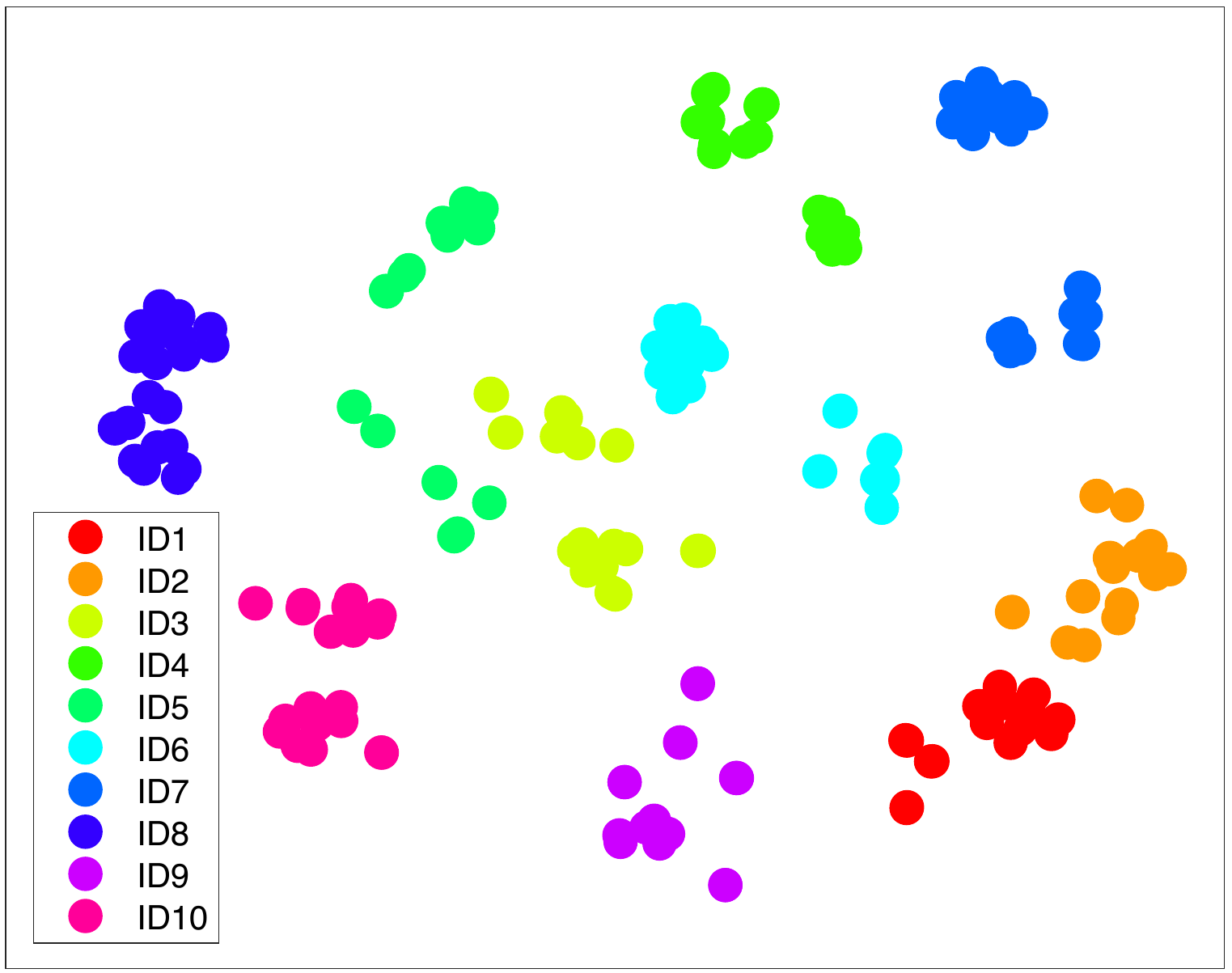}&
 \includegraphics[width=4cm,height=3cm]{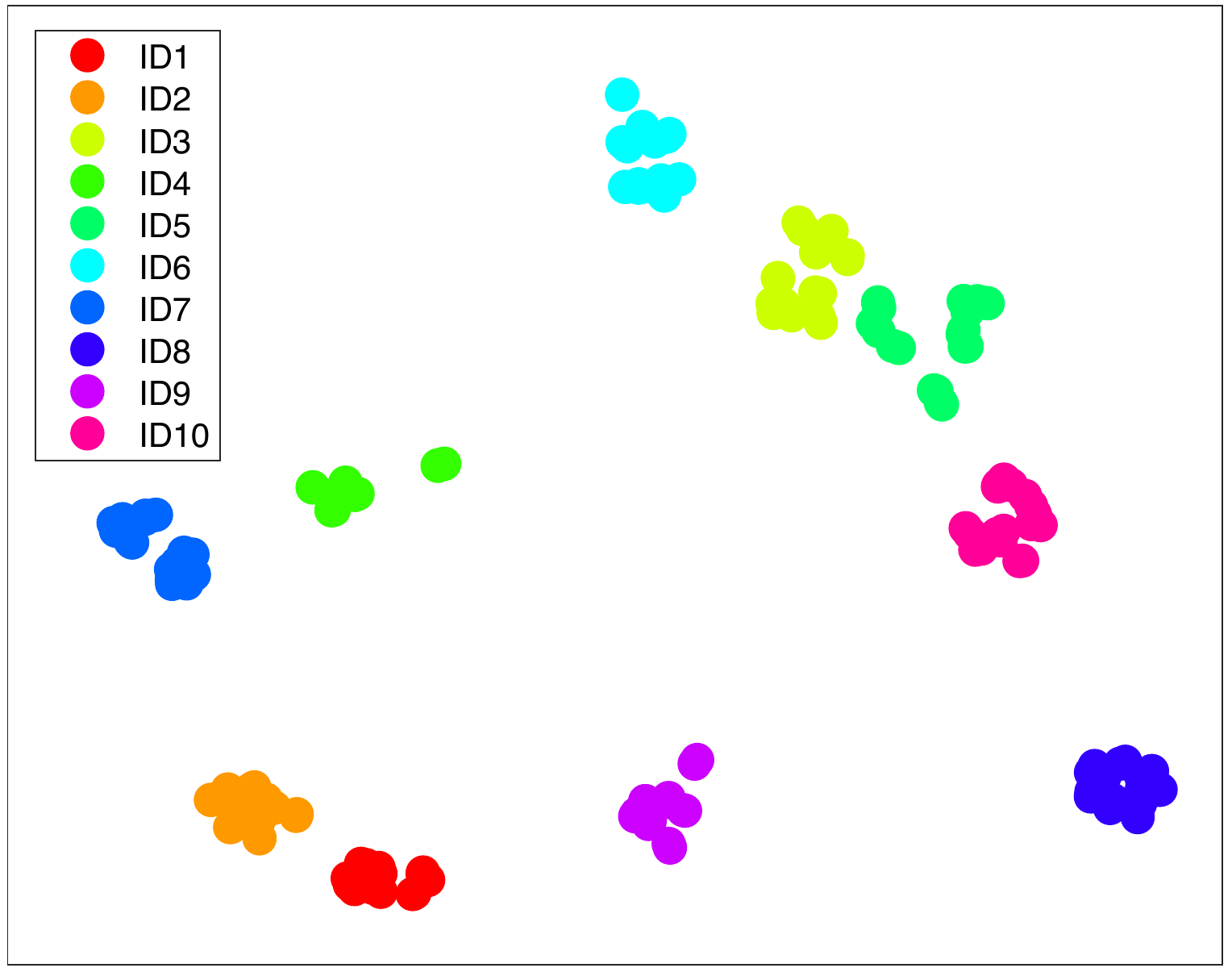}\\
 \footnotesize{(c) Softmax(ResNet50,2048-D)} & \footnotesize{(d) ours(ResNet50,2048-D)}
 \end{tabular}
 \vspace{.4em}
 \caption{2D tSNE visualization of the learned pedestrian features on the CUHK03 test datasets for ten persons. All the features are learned based on the ResNet50  with different loss function. a)BGCL,b)BGTL,(c)BGTL and d)ours refer to the batch-global contrastive loss, batch-global triplet loss, softmax loss and the proposed joint learning method. Different persons are color coded.}
 \label{t-SNEVisualization}
 \end{figure}

 \textbf{The t-SNE visualization.} In Figure~\ref{t-SNEVisualization}, we extract the deep features on CUHK03 test dataset with the ResNet50 model trained by four different loss functions, and then visualize them in 2D using the t-SNE algorithm~\cite{maaten2008visualizing}. We can clearly see that our method can help making the learned features with better intra-personal compactness and inter-personal separability as compared to the corresponding baseline models.


\section{Conclusion}~\label{SectionConclusion}

In this paper, we propose to formulate the traditional contrastive and triplet loss into a structured graph Laplacian form by construction one complete graph. This step enables the proposed methods can take full advantages of the cluster structure information in the training batches. When combining the proposed graph Laplacian and the softmax loss for joint training, our method can obtain discriminative deep features with inter-personal dispersion and intra-personal compactness.
In the future, we tend to explore more distance relationships in the graph Laplacian framework, and then extend our method to other visual recognition tasks.


{
\bibliographystyle{ieee}
\bibliography{egbib}
}

\end{document}